\documentclass[trsc,nonblindrev]{informs3} 

\OneAndAHalfSpacedXI 

\renewcommand{\theARTICLETOP}{}
\renewcommand{\theRRHFirstLine}{}
\renewcommand{\theRRHSecondLine}{}
\renewcommand{\theLRHFirstLine}{}
\renewcommand{\theLRHSecondLine}{}
\usepackage{soul}
\usepackage{natbib}
 \bibpunct[, ]{(}{)}{,}{a}{}{,}%
 \def\bibfont{\small}%
 \usepackage{mathtools}
\usepackage{mathpazo}
\usepackage{amsmath}
\usepackage{mathrsfs}
\usepackage{amssymb}
\usepackage{multirow}
\usepackage{longtable}
\usepackage{graphicx}
\usepackage{enumerate}
\usepackage{graphicx} 
\usepackage{epstopdf}
\usepackage{svg}
\usepackage{subfigure}
\usepackage{algorithm,algorithmic,longtable}
\usepackage{graphicx}
\usepackage{indentfirst}
\usepackage[colorlinks=true, linkcolor=blue, citecolor=blue, urlcolor=blue]{hyperref}

\usepackage{booktabs}
\usepackage{array, caption, threeparttable,arydshln}
\usepackage{caption}
\DeclareCaptionLabelSeparator{twospace}{\ ~\ ~}   
\newtheorem{thm}{Theorem}

\usepackage{empheq}
\usepackage{cases}

\allowdisplaybreaks[1]
\usepackage{longtable}
\usepackage{multicol} 
\renewcommand{\algorithmicrequire}{\textbf{Input:}}
\renewcommand{\algorithmicensure}{\textbf{Output:}}

\makeatletter
\newcommand{\mathleft}{\@fleqntrue\@mathmargin\parindent}
\newcommand{\mathcenter}{\@fleqnfalse}
\makeatother
\usepackage{etoolbox}
\newcommand{\zerodisplayskips}{%
\setlength{\abovedisplayskip}{7pt}%
\setlength{\belowdisplayskip}{6pt}%
\setlength{\abovedisplayshortskip}{7pt}%
}
\appto{\normalsize}{\zerodisplayskips}
\appto{\small}{\zerodisplayskips}
\appto{\footnotesize}{\zerodisplayskips}

\allowdisplaybreaks \allowdisplaybreaks[1]

\EquationsNumberedThrough    

\MANUSCRIPTNO{} 

\begin{document}

\TITLE{Towards Full-Scenario Safety Evaluation of Automated Vehicles: A Volume-Based Method}

\ARTICLEAUTHORS{%

\AUTHOR{Hang Zhou$^1$, Chengyuan Ma$^1$, Shiyu Shen$^2$, Zhaohui Liang$^1$, Xiaopeng Li\footnote{Corresponding author. Email: xli2485@wisc.edu}$^1$}
\AFF{$^1$ Department of Civil and Environmental Engineering, University of Wisconsin-Madison}
\AFF{$^2$ Department of Civil and Environmental Engineering, University of Illinois Urbana-Champaign}
}

\ABSTRACT{
With the rapid development of automated vehicles (AVs) in recent years, commercially available AVs are increasingly demonstrating high-level automation capabilities. However, most existing AV safety evaluation methods are primarily designed for simple maneuvers such as car-following and lane-changing. While suitable for basic tests, these methods are insufficient for assessing high-level automation functions deployed in more complex environments. First, these methods typically use crash rate as the evaluation metric, whose accuracy heavily depends on the quality and completeness of naturalistic driving environment data used to estimate scenario probabilities. Such data is often difficult and expensive to collect. Second, when applied to diverse scenarios, these methods suffer from the curse of dimensionality, making large-scale evaluation computationally intractable.
To address these challenges, this paper proposes a novel framework for full-scenario AV safety evaluation. A unified model is first introduced to standardize the representation of diverse driving scenarios. This modeling approach constrains the dimension of most scenarios to a regular highway setting with three lanes and six surrounding background vehicles, significantly reducing dimensionality. To further avoid the limitations of probability-based method, we propose a volume-based evaluation method that quantifies the proportion of risky scenarios within the entire scenario space. For car-following scenarios, we prove that the set of safe scenarios is convex under specific settings, enabling exact volume computation. Experimental results validate the effectiveness of the proposed volume-based method using both AV behavior models from existing literature and six production AV models calibrated from field-test trajectory data in the Ultra-AV dataset. Code and data will be made publicly available upon acceptance of this paper.
}

\KEYWORDS{Automated Vehicle; Safety Evaluation; Full-scenario Autonomous Driving}
\maketitle

\section{Introduction}

Automated vehicle (AV) technology has advanced rapidly over the past decade. During this period, an increasing number of commercially available vehicles have been equipped with self-driving capabilities \citep{li2022trade}. Although most AVs currently on the market remain at Level 2 automation, several leading AV manufacturers, such as Tesla, Waymo, Huawei, Xpeng, and Li Auto, have recently announced the successful implementation of the "parking-to-parking" feature. This feature enables fully autonomous driving from a parking space at the origin to a parking space at the destination \citep{huawei2025parking}. However, the deployment of these features in real-world environments presents significant safety challenges \citep{ding2023survey,coppola2023assessing}. Rigorous and comprehensive safety testing is essential to ensure the reliability, robustness, and public acceptance of both current and future AVs \citep{moody2020public,ma2024evolving}.

Existing literature on AV safety evaluation primarily focuses on relatively simple and specific driving scenarios. For example, \citet{zhao2017accelerated} and \citet{ma2023trajectory} evaluated the adaptive cruise control function in the car-following scenario. \citet{zhao2016accelerated,feng2020testing3,feng2021intelligent}, and \citet{zhou2025quantile} extended the testing scenarios to include multiple lanes and lane-changing behaviors. \citet{arvin2020safety,tang2021systematic} and \citet{song2022intersection} examined AV behavior at intersections. \citet{feng2020testing2} conducted a series of case studies to validate the general framework proposed in \citet{feng2020testing}, covering car-following, cut-in, and highway exit scenarios. In addition, \citet{feng2023dense} tested a proposed scenario generation method in a roundabout setting. The primary reason for the literature to focus on relatively simple scenarios is that commercial AVs over the past several years have mostly remained at low levels of automation. The limited availability of AVs with high levels of automation has resulted in less urgent demand for evaluation methods targeting complex environments and a shortage of relevant field test data necessary for the development and calibration of AV behavior models. However, as high-level AVs begin to emerge, existing evaluation methods designed and tested on car-following and lane-changing scenarios are no longer sufficient to meet the growing safety testing needs.

\begin{figure}
    \centering
    \includegraphics[width=0.7\linewidth]{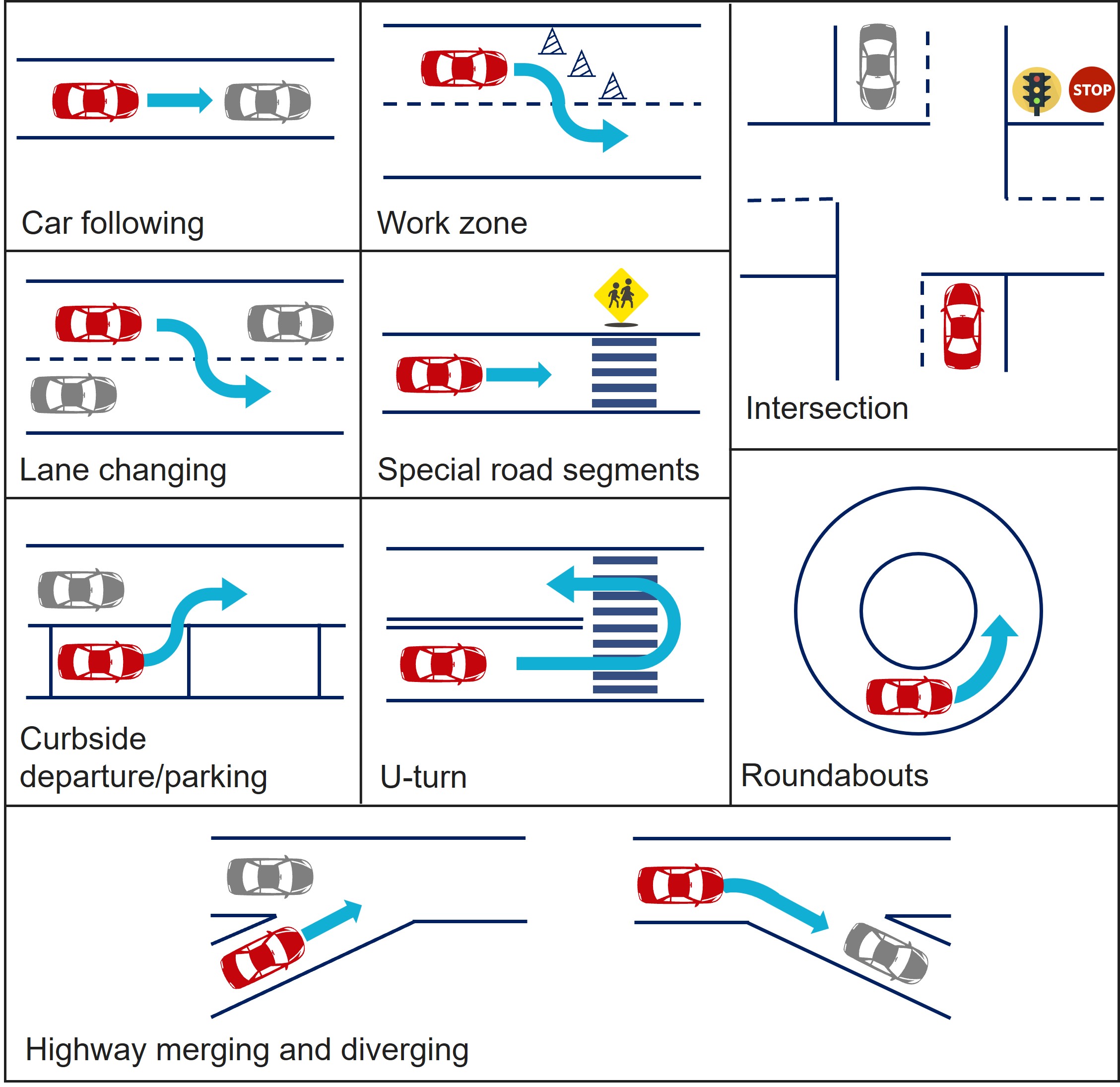}
    \caption{Examples of different driving scenarios.}
    \label{fig:scenario}
\end{figure}

Safety testing for full-scenario autonomous driving presents significant challenges, primarily due to the complexity and variability of driving environments. Moreover, the absence of a unified scenario model hinders the establishment of a consistent understanding of the role that scenarios play in the AV development process \citep{ma2021traffic,ren2022survey}. Figure~\ref{fig:scenario} illustrates several representative types of driving scenarios encountered in real-world settings. Each scenario involves different types of agents that must be accurately modeled. For example, road segment scenarios require consideration of pedestrians and sidewalks, while curbside scenarios necessitate the representation of various types of parking spaces. The presence of numerous agents across diverse scenarios results in a high-dimensional state space, which significantly reduces the efficiency of AV safety testing. This phenomenon is commonly referred to as the curse of dimensionality \citep{feng2023dense}. Consequently, establishing a unified scenario model that covers all scenario types with an acceptable dimension is necessary for AV evaluation.

From the perspective of evaluation method, existing literature mainly focuses on probability-based methods, such as estimating the crash rate of AVs \citep{zhao2017accelerated,feng2021intelligent,feng2023dense}. These approaches typically extract the behavior distribution of agents at each state from large-scale naturalistic driving environment (NDE) data, and then compute the probability of the entire driving scenario \citep{roesener2016scenario}. Although these methods can be integrated into our structured scenario model, they are difficult to implement in full-scenario evaluation. This is because the probability-based method heavily depends on the completeness and accuracy of the behavior distributions extracted from NDE data, which are challenging to collect and process for all possible driving behaviors. If the NDE dataset is not sufficiently large, the extracted behavior distributions may be inaccurate, leading to unreliable estimates of the crash rate. Furthermore, due to the low probability of dangerous behaviors, sampling methods based on probability often encounter the challenge known as the curse of rarity \citep{liu2024curse}, which significantly increases the evaluation time.

To address the research gaps mentioned above, this study proposes a novel framework for full-scenario AV safety evaluation. First, to meet the testing requirements of full-scenario autonomous driving, we develop a unified scenario model that represents a wide range of driving scenarios with reduced dimensional complexity. In this model, the driving process is described as a dynamic process, and the definitions of different scenario types are standardized. The overall dimensionality of the model is comparable to that of a typical highway scenario with three lanes, making it compatible with existing evaluation methods.
Second, to overcome the limitations of probability-based evaluation methods caused by their reliance on NDE data, we propose a volume-based evaluation approach. This method quantifies the proportion of dangerous scenarios within the entire space of possible driving scenarios. Theoretical analysis shows that when evaluating AVs with strong safety performance, such as those with high levels of automation, the volume-based and probability-based methods yield approximately equivalent results in terms of safety comparison.
To support this evaluation, we develop a Monte Carlo algorithm for the unified scenario model. Furthermore, for car-following scenarios, we prove that the safe scenario set forms a convex polytope in a high-dimensional space. This convexity property allows for the exact computation of scenario volume and enables the use of sampling-based algorithms designed for convex sets.
Experimental results demonstrate that the proposed method can efficiently evaluate safety within the unified scenario framework. Using the developed evaluation pipeline, we perform a safety ranking of several commercial AVs based on their adaptive cruise control function. The results validate the effectiveness and practicality of the proposed framework.
The main contributions of this study are summarized as follows:

\begin{itemize}
    \item We propose a unified scenario model that generalizes the vast variety of traffic situations into an onboard-view representation limited to at most three lanes (the ego lane and its adjacent lanes) and no more than six surrounding vehicles. This abstraction significantly reduces the policy dimensionality while preserving critical driving complexity.

    \item We design a volume-based evaluation method combined with a Monte Carlo algorithm for full-scenario AV safety analysis. By comparing the relative frequency of potential accidents across different vehicles, rather than directly estimating and comparing absolute crash rates, this method avoids the impractical need to accurately model the probability of various critical states.

    \item Theoretical analysis shows that for highly automated AVs, the volume-based and probability-based methods yield approximately equivalent safety ranking results. In addition, we prove that for car-following scenarios, the set of safe scenarios forms a convex polytope, which allows for analytical computation in volume-based evaluation.
\end{itemize}

The remainder of this paper is organized as follows. Section \ref{sec:frame} builds the volume-based evaluation framework for full-scenario AVs. Section \ref{sec:alg} proposes the algorithms to conduct volume-based evaluation. Section \ref{sec:exp} shows the experimental results to validate the proposed evaluation framework and discussions. Section \ref{sec:con} concludes this paper and provides future work.

\section{An Evaluation Framework for Full-scenario Autonomous Driving}
\label{sec:frame}

In this section, we present our evaluation framework, which includes a unified model for different scenario types and a volume-based evaluation approach. We begin by introducing the basic concepts and notations in Section~\ref{sec:concept}, followed by the detailed modeling framework in Section~\ref{sec:model}. The volume-based evaluation approach is described in Section~\ref{sec:volume}.

\subsection{Basic Concepts and Notations}
\label{sec:concept}

Although a unified scenario model is currently lacking, prior work has provided valuable insights into the fundamental structure of driving scenarios \citep{geyer2014concept,zhu2019review,ren2022survey}. Based on these studies, we identify the core elements of a scenario in the context of autonomous driving. Therefore, before introducing our proposed scenario model, we first define the key concepts used within the model.

\begin{figure}
    \centering
    \includegraphics[width=1\linewidth]{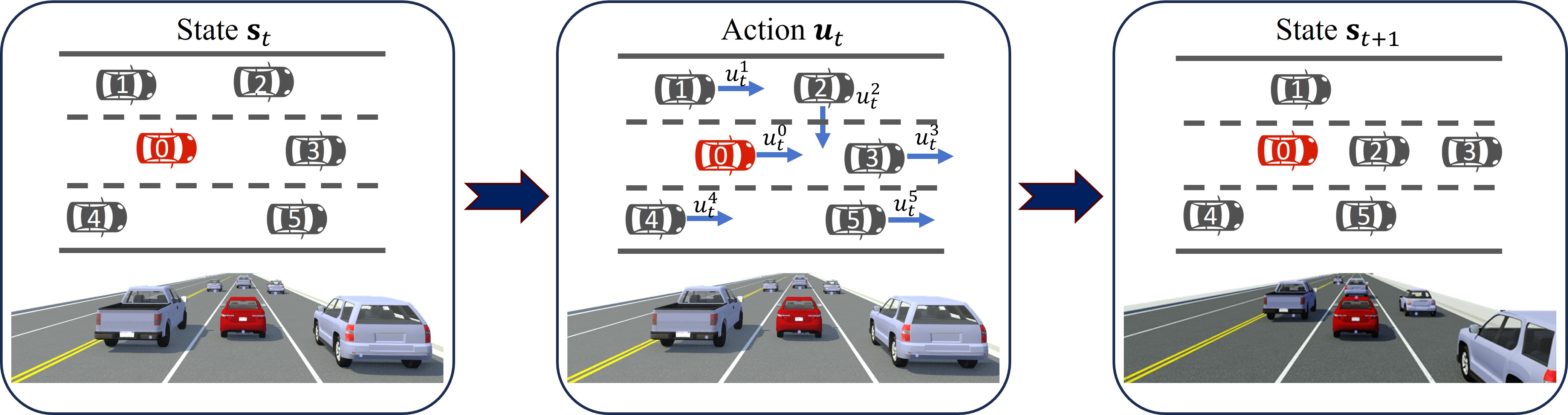}
    \caption{Illustration of the state transition process, where the red vehicle represents the tested AV and black vehicles represent BVs.}
    \label{fig:state}
\end{figure}

We define the driving scenario as a dynamic process over a discrete-time horizon \(\mathcal{T}=\{0,1,\dots, T\} \), where \(T\in\mathbb{N}\) represents the total number of time steps, and \(\Delta t\) represents the time interval between two consecutive steps. Denote $\mathcal{I}=\{0,1,\dots,I\}$ as the index set of agents involved in the scenario, where index $0$ represents the AV and $1,\dots,I$ represent other agents such as the background vehicles (BVs). Denote \( \mathcal{S}=\{\mathbf{s}_t\}_{t\in \mathcal{T}} \) as the set of \textbf{states} in the time horizon, where $\mathbf{s}_t=[\mathbf{s}^0_t,\mathbf{s}^1_t,...,\mathbf{s}^I_t]$ is a vector including the information $\mathbf{s}^i_t$ for agent $i\in \mathcal{I}$ at time $t\in \mathcal{T}$, such as the position and speed. \( \mathcal{U}=\{\mathbf{u}_t\}_{t\in \mathcal{T}} \) denotes the \textbf{actions} in the time horizon, where $\mathbf{u}_t=[u^0_t,u^1_t,...,u^I_t]$ is a vector including the action $u^i_t$ for agent $i\in \mathcal{I}$ at time $t\in \mathcal{T}$, such as the longitudinal acceleration and the lateral action (i.e., left lane change, straight, and right lane change). As illustrated in Figure~\ref{fig:state}, the entire dynamic process starts from \(t = 0\) and evolves iteratively. At each state \(\mathbf{s}_t\), the next state \(\mathbf{s}_{t+1}\) is determined based on the action \(\mathbf{u}_t\). Thus, a \textbf{testing scenario} can be defined by the initial state of the system and the actions of the BVs, which is denoted as \( \mathbf{x} = (\mathbf{s}_0, \mathbf{u}_0, \mathbf{u}_1, \ldots, \mathbf{u}_{T-1})\). Once the behavior model of the tested-AV and the testing scenario are specified, the entire dynamic process can be determined, including the states and actions of all agents at each time step. We refer to this determined process as a \textbf{driving scenario}.

To help readers better understand our model, we use a car-following scenario with a lead vehicle (LV) and a following vehicle (FV) in a single lane as an illustrative example. All definitions below refer to a specific testing scenario \(\mathbf{x}\); for simplicity, we omit \(\mathbf{x}\) in the notation. In this type of scenario, the agents are the LV and FV, denoted by superscripts \(\mathrm{l}\) and \(\mathrm{f}\), respectively. The state at each time step \(t \in \mathcal{T}\) is represented by the speeds of the LV and FV, denoted as \(v^{\mathrm{l}}_t\) and \(v^{\mathrm{f}}_t\), and the spatial gap between them, denoted as \(d_t\). The actions are defined as the acceleration of the LV and FV, which are denoted as \(a^{\mathrm{l}}_t\) and \(a^{\mathrm{f}}_t\). A testing scenario can be specified by the initial spatial gap \(d_0\) and the speed profile of the LV \(\{v^{\mathrm{l}}_t\}_{k \in \mathcal{T}}\). In this setting, the behavior model is a car-following model, which can be formulated as:
\begin{align}
   a^{\mathrm{f}}_t = f(v^{\mathrm{l}}_t, v^{\mathrm{f}}_t, d_t, v^{\mathrm{l}}_{t-1}, v^{\mathrm{f}}_{t-1}, d_{t-1},...,v^{\mathrm{f}}_0, v^{\mathrm{f}}_0, s_0).
\end{align}

Once FV's acceleration $a^{\mathrm{f}}_t$ at time $t$ is obtained, the state for the next time step $t+1$ can be updated by
\begin{align}
   & v^{\mathrm{f}}_{t+1} = v^{\mathrm{f}}_t + a^{\mathrm{f}}_t\Delta t,\\
   & d_{t+1}=d_{t}+\Delta t v^{\mathrm{l}}_{t}+\frac{\Delta t^2}{2} a^{\mathrm{l}}_{t}-\Delta t v^{\mathrm{f}}_{t}-\frac{\Delta t^2}{2} a^{\mathrm{f}}_{t}.
\end{align}
Considering the vehicle dynamics and road condition constraints, such as positive initial gap and vehicle speed \citep{zhou2024unified}, we restrict the scenario space to a bounded region $\Omega \subseteq \mathbb{R}^M$, where the dimension $M = 3 + T$.

\subsection{A Unified Model for All Scenario Types}
\label{sec:model}

In the previous section, we proposed the basic concepts of our modeling method, i.e., the agent, state, action, testing scenario, and driving scenario, and provided an example for the car-following scenario. However, if we aim to model other types of scenarios illustrated in Figure~\ref{fig:scenario}, special designs for agents, states, and actions (i.e., \(\mathcal{I}, \mathcal{S}, \mathcal{U}\)) are required. For instance, in pedestrian scenarios, pedestrians need to be considered as agents; in merging scenarios, the lane position should be included in the agent's state; and in U-turn scenarios, the AV's action should account for turning around. Since there are numerous types of driving scenarios, designing specific models for each scenario would require substantial expert knowledge and result in a high overall dimensionality. in this section we define the details of the unified model to make the dimension less or equal than a highway scenario with three lanes.

\subsubsection{State.}
We first consider the definition of states. The positions of all agents in the driving process can be described based on the lanes. For some especial scenarios such as highway merging and diverging, work zones, or curbside parking, the corresponding special lane and parking position can be defined as lanes of finite length, or static agents can be placed at certain positions within the lane. Thus the lane ID for the agent is an important feature. In a specific lane, we only consider the agent's longitudinal position, i.e., the position along the lane direction, and do not consider its lateral position. This is because a longitudinal overlap between the AV and the agent in the same lane is considered as collision, regardless of the agent's lateral position. For instance, in pedestrian crossing scenarios, it is unnecessary to model the exact lateral position of the pedestrian; it suffices to treat the pedestrian as an agent with zero longitudinal speed performing a lane-changing behavior. Therefore, the state of an agent can be described using four variables: lane ID, distance, longitudinal speed, and lane-changing time. The distance refers to the longitudinal travel distance relative to the initial position when entering the lane, and the lane-changing time refers to the time spent since the start of the lane-changing behavior.

\subsubsection{Action.}
Based on the state definition, the agent's action can be defined as longitudinal and lateral behaviors, i.e., longitudinal acceleration and lane-changing behavior. Other special driving behaviors can be incorporated into these two categories. For example, lane changes at ramp entrances can be treated as lane changes to the main lane; U-turns can be treated as lane changes to the opposite lane; starting from curbside parking can be treated as lane changes to the driving lane; yielding to pedestrians or waiting at traffic lights can be modeled as decelerating to a stop; pedestrian crossing can be modeled as a lane-changing behavior.

\subsubsection{Agent.}
Agents can be defined as any entity present on the lane. In most cases, we consider only the six surrounding agents closest to the AV, as illustrated by the vehicles marked with red circles in Figure~\ref{fig:horizon}. This is based on the observation that vehicle behavior is primarily influenced by the nearest vehicle in each direction. In addition to common traffic participants, this also includes obstacles in work zones and some virtual agents, such as assuming a virtual agent exists during a red light at an intersection. The differences between agent types are limited to their physical constraints, such as bounds on acceleration, speed, and lane-changing time. For instance, obstacles can only remain static, and pedestrians have zero longitudinal speed.

\subsubsection{Collision.}
\begin{figure}
    \centering
    \includegraphics[width=0.5\linewidth]{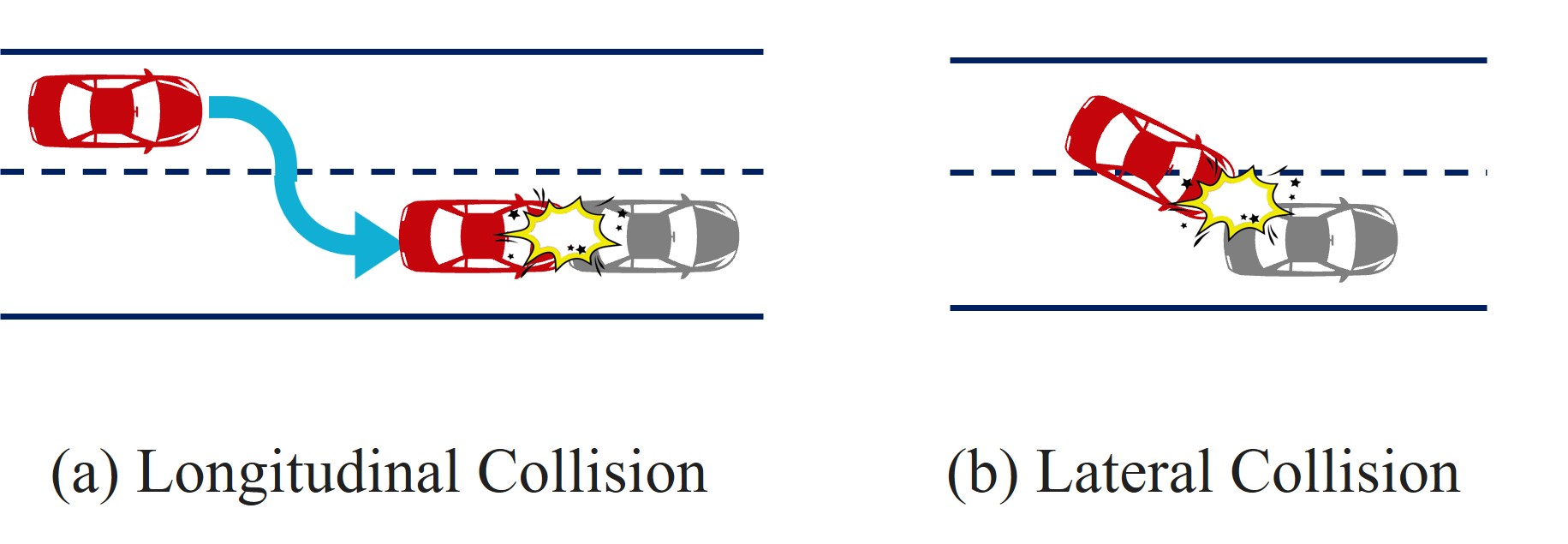}
    \caption{Illustration of the longitudinal and lateral collisions.}
    \label{fig:collision}
\end{figure}

In AV safety evaluation, the primary focus is on the dangerous scenarios, particularly the collisions. The National Highway Traffic Safety Administration (NHTSA) categorizes vehicle-to-vehicle collision types into multiple classes \citep{ulfarsson2006factors}. These collision types can be simplified into two categories based on the types of actions that lead to collisions: \textbf{longitudinal collisions} and \textbf{lateral collisions}. Longitudinal collisions refer to rear-end collisions occurring within the same lane, as illustrated in Figure~\ref{fig:collision}. These are not limited to long-term car-following behavior but often occur during the short car-following phase after a lateral maneuver, where an improper lane change results in a small spatial gap or a large speed difference between vehicles. Lateral collisions refer to side or corner collisions caused when the ego vehicle and another vehicle have partial longitudinal overlap while performing lateral behaviors, as shown in Figure~\ref{fig:collision}. 

\subsubsection{Time Horizon.}
\begin{figure}
    \centering
    \includegraphics[width=0.75\linewidth]{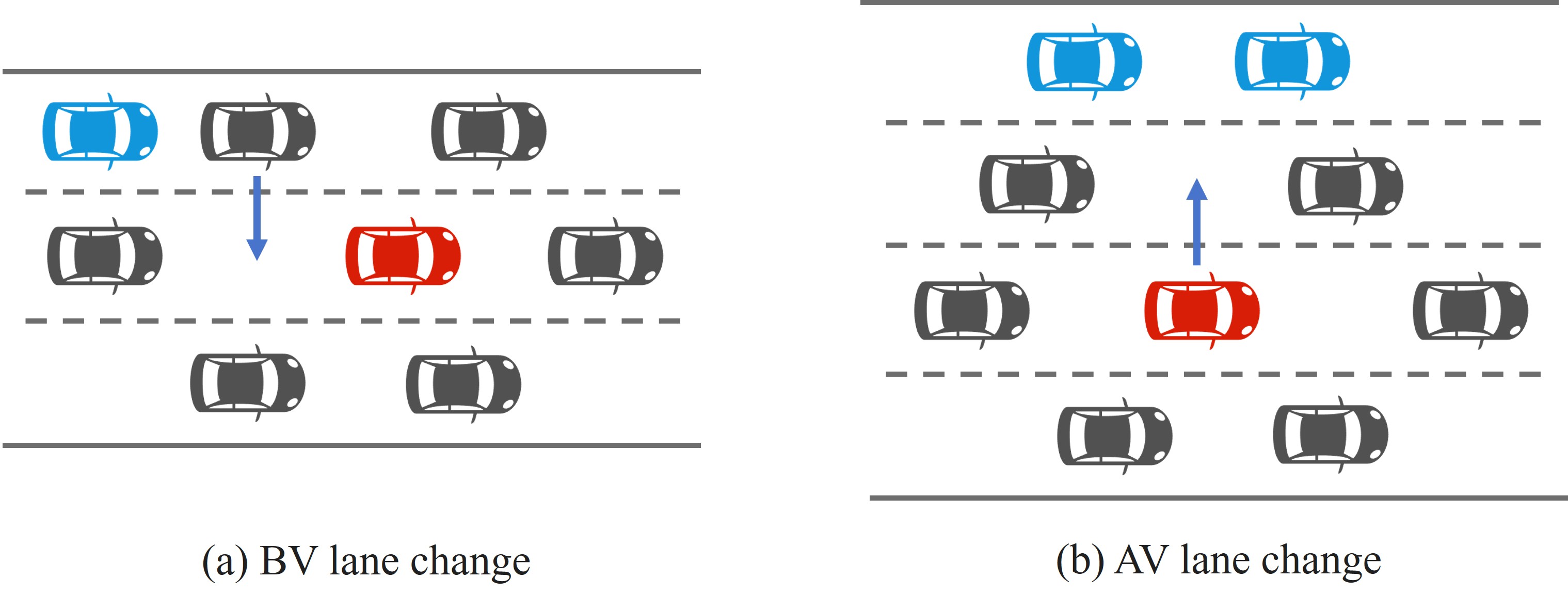}
    \caption{Examples of early termination of a driving scenario: (a) agent set update due to BV lane change; (b) surrounding agent change caused by AV lane change.}
    \label{fig:horizon}
\end{figure}

Although we define the time horizon of a driving scenario as \(T\), in practice, the scenario may terminate earlier under certain conditions. These include: (1) a collision occurs; (2) the set of six surrounding agents changes; or (3) the AV performs a lane-change maneuver. The latter two termination conditions are intentionally designed to limit the dimensionality of the system. If such changes are included within a single driving scenario, the model must account for additional agents and lane information, significantly increasing complexity. For example, as illustrated in Figure~\ref{fig:horizon}(a), the AV initially considers six surrounding agents, marked by a gray color. However, if the rear-left BV executes a lane change, the AV must then account for the behavior of the newly involved BV shown in blue. This increases the number of agents that influence the AV's decision. To prevent the unbounded growth of agents in the system, we split the complete trajectory into two separate driving scenarios: the trajectory after the agent update is treated as a new scenario initialized with the updated agent set. Similarly, Figure~\ref{fig:horizon}(b) illustrates a case where a lane change by the AV causes a shift in the surrounding agents, which would also increase the system complexity if modeled continuously. Therefore, both agent set changes and AV lane changes are treated as termination conditions for a driving scenario.

\subsubsection{Additional Setting.}
Based on the proposed model, additional designs can be incorporated depending on the granularity of the evaluation task. For example, some studies introduce further assumptions on the agent's actions, modeling the process as a stochastic process or a Markov process \citep{feng2021intelligent,feng2023dense}. These assumptions can be seamlessly integrated into the proposed framework.

\subsubsection{Examples.}

\begin{figure}
    \centering
    \includegraphics[width=0.6\linewidth]{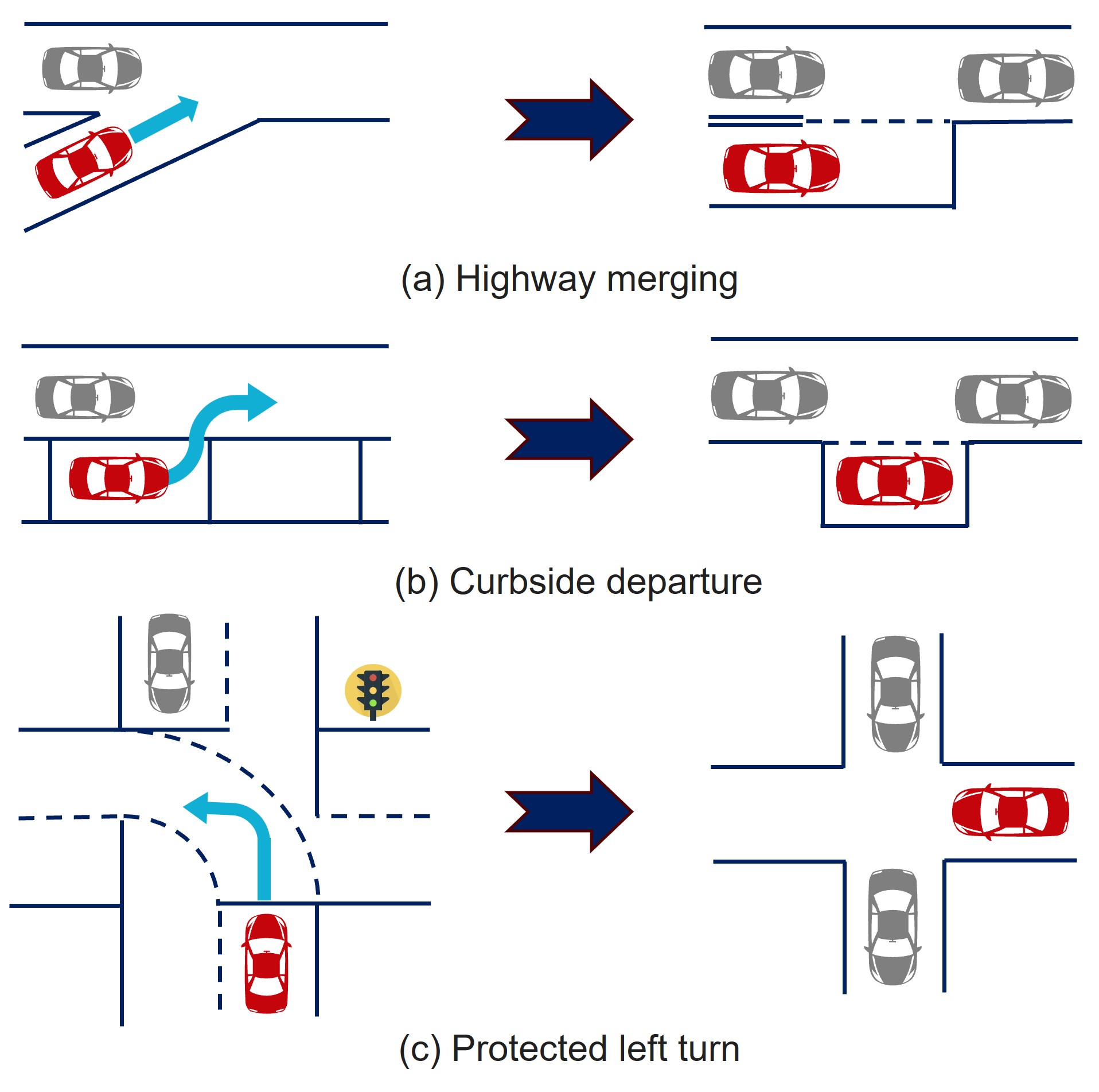}
    \caption{Examples of the unified scenario model.}
    \label{fig:examples}
\end{figure}

Figure~\ref{fig:examples} presents several examples of how different scenario types can be transformed into the unified scenario model. Figure~\ref{fig:examples}(a) illustrates a highway merging scenario, which can be simplified into a two-lane, two-agent setting, where the merging ramp is modeled as a lane segment with limited length. Figure~\ref{fig:examples}(b) shows a curbside departure scenario, which can be similarly transformed into a structure analogous to that in (a). Figure~\ref{fig:examples}(c) depicts a more complex protected left-turn scenario at a signalized intersection. In this case, the left-turn lane can be interpreted as a lateral lane, while the opposing through lane is treated as a longitudinal lane. The AV’s core decision-making task in this setting involves performing gap acceptance between agents on the two longitudinal lanes.

These scenario transformations are generally consistent with the unified model's definitions. Although minor customizations are needed in some special cases, such as signalized intersections, these scenarios can still be represented within the unified framework using a relatively low-dimensional formulation.

\subsection{Volume-based Evaluation Method}
\label{sec:volume}

In the literature, AV safety evaluation is often based on probability, such as estimating the crash rate. If the driving process is assumed to follow a Markov property \citep{feng2021intelligent,feng2023dense}, the probability of a driving scenario $x$ can be decomposed as
\begin{align}
    P(\mathbf{x}) = P(\mathbf{s}_0) \times \prod_{t=0}^{T} P(\mathbf{u}_t \mid \mathbf{s}_t), \label{eq:pro}
\end{align}
where \(P(\mathbf{u}_t \mid \mathbf{s}_t)\) denotes the probability that agents take actions \(\mathbf{u}_t\) under state \(\mathbf{s}_t\), which can be derived from agents' behavior distributions. Then, the crash rate can be defined as the sum of the probability of the crash driving scenario. Our structured scenario modeling framework can easily incorporate this probability-based method. However, such analysis heavily relies on the completeness and accuracy of the behavior distributions extracted from NDE data, which are difficult to collect for all scenario types. To address these limitations, we propose a volume-based evaluation method. This approach does not consider the probability of scenarios, so it does not require collecting large-scale NDE data to model or estimate the behavior distributions of BVs, nor does it rely on extensive sampling to ensure rare scenarios are captured. Before introducing the volume-based method, we first define the dangerous scenario according to the safety metrics.

The safety level of a driving scenario can be evaluated using surrogate safety metrics. Denote \(\mathrm{SM}\) as a surrogate safety metric. For simplicity, we assume that smaller values of the safety metric indicate more dangerous conditions. In this study, we use a widely adopted safety metric: Time to Collision (TTC) \citep{wang2021review}. TTC measures the time required for the FV to reach the current position of the LV, assuming both vehicles maintain their current speeds. A lower TTC indicates a higher risk of collision. In a car-following scenario, TTC is calculated as follows:
\begin{align}
  \mathrm{TTC}_t =
  \begin{cases}
     \frac{d_t-l}{v^{\mathrm{f}}_t-v^{\mathrm{l}}_t}, 
     & v^{\mathrm{f}}_t-v^{\mathrm{l}}_t > 0,\\[6pt]
     +\infty,
     & \text{otherwise}.
  \end{cases} \label{eq:ttc}
\end{align}
Where $l$ is the vehicle length. It is clear that \(\mathrm{TTC} = 0\) indicates a collision. Therefore, in multi-lane scenarios, lateral collisions are also denoted by \(\mathrm{SM} = 0\). The overall safety level for a scenario can be measured as the minimum TTC among the entire time horizon $\min_{t\in\mathcal{T}} \mathrm{TTC}_t$. Note that other safety metrics, such as Deceleration Rate to Avoid Collision and time headway \citep{wang2021review}, can also be used for evaluation.

Based on the safety metric, we define a dangerous scenario by the minimum value of \(\mathrm{SM}(\mathbf{x})\) within the scenario to a predefined threshold \(\eta\). Specifically, scenario \(\mathbf{x}\) is considered dangerous if \(\mathrm{SM}(\mathbf{x}) \leq \eta\); otherwise, it is classified as a safe scenario. The sets of dangerous and safe scenarios are defined as follows:
\begin{align}
   \mathcal{D} & = \{\mathbf{x}\in \Omega \mid \mathrm{SM}(\mathbf{x}) \leq \eta \} \label{eq:dangerous}\\
   \mathcal{S} & = \Omega \setminus \mathcal{D} = \{\mathbf{x} \in \Omega \mid \mathrm{SM}(\mathbf{x}) > \eta\}.
\end{align}

\begin{figure}
    \centering
    \includegraphics[width=0.75\linewidth]{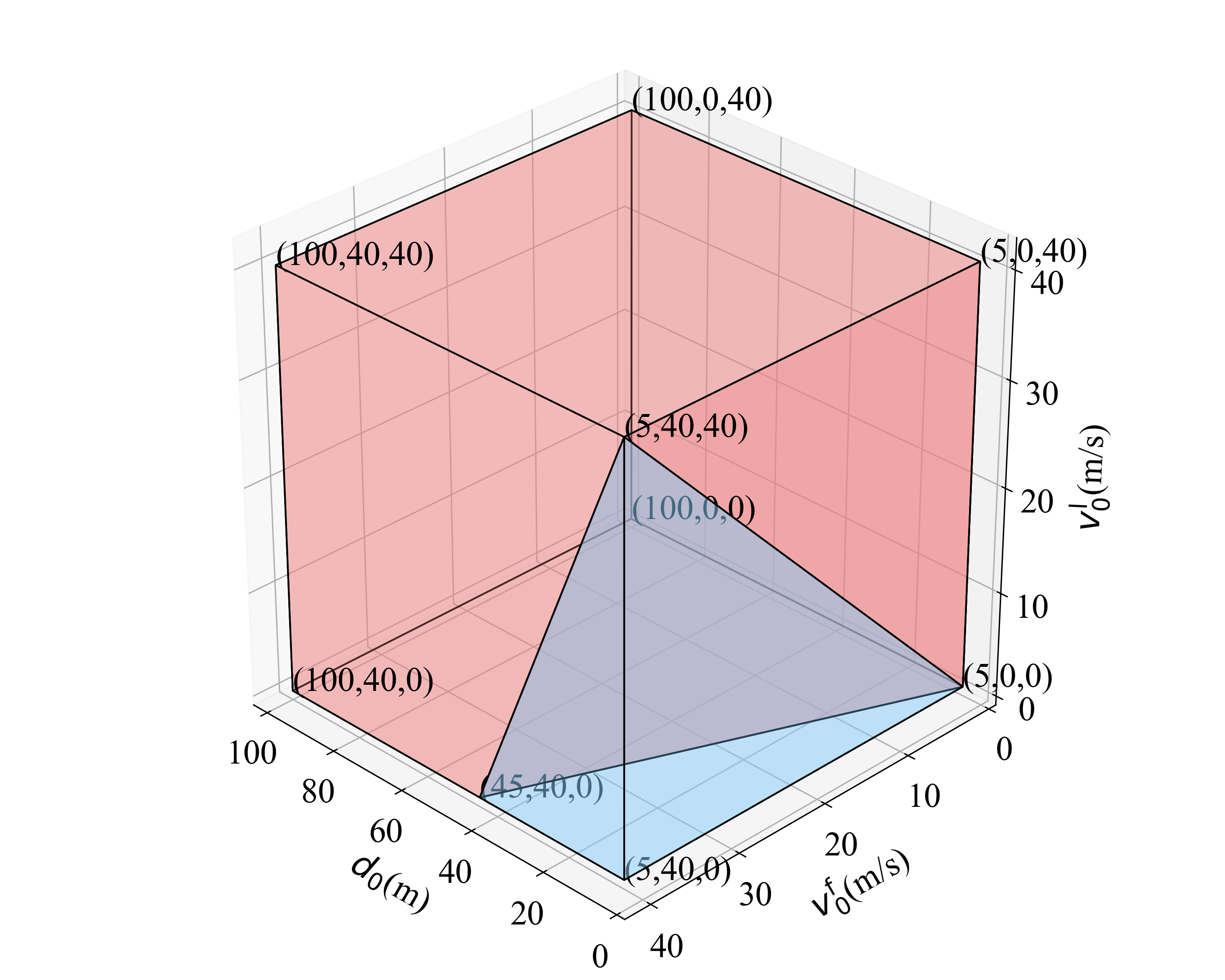}
    \caption{Example of safe (red) and dangerous (blue) scenario sets at \( T = 0 \) in a car-following scenario.}
    \label{fig:poly}
\end{figure}

Using algorithms that will be described in Section~\ref{sec:alg}, we can estimate the volume of the dangerous scenario set \(\mathcal{D}\), denoted as \(\mathrm{vol}(\mathcal{D})\). This volume measure can serve as an indicator of the safety performance of the tested AV. Figure~\ref{fig:poly} illustrates an example of the spatial distribution of safe and dangerous scenario sets for a car-following scenario with \(T = 0\). For visualization purposes, the vehicle length is set to \(l = 0\), and the threshold for the dangerous scenario is defined as \(\eta = 1\)~s based on the TTC safety metric. Although a scenario with \(T = 0\) is not suitable for evaluating AV behavior, this example provides an intuitive visualization of how safe and dangerous scenarios are distributed in the state space. In this figure, each point in the coordinate space represents a unique driving scenario. The safe scenario set is shown in red, while the dangerous scenario set is shown in blue. The boundary plane between the red and blue regions corresponds to scenarios with \(\mathrm{TTC} = 1\)~s, and is defined by the three points \((45, 40, 0)\), \((5, 40, 40)\), and \((5, 0, 0)\). One might argue that the volume itself may not carry explicit physical meaning, as it includes many dangerous scenarios that are inherently unavoidable. However, this measure still enables fair comparisons across different AV systems. In the following analysis, we illustrate how the volume-based method differs from the probability-based method, and under what conditions their conclusions align.

\begin{thm}
    \label{thm:equivalent}
    When comparing the safety performance of highly automated AVs, the volume-based and probability-based methods yield approximately equivalent results.
\end{thm}

\noindent \textit{Proof of Theorem \ref{thm:equivalent}:}  
We first analyze the probability-based method. According to Equation~\eqref{eq:pro}, the probability of a driving scenario \(\mathbf{x}\) can be estimated. Based on the definition of the dangerous scenario set in Equation~\eqref{eq:dangerous}, the crash rate can be expressed as:
\begin{align}
   P(\mathcal{D}) = \int_{\mathbf{x}\in \Omega} P(\mathbf{x}) \cdot \mathbf{1}\{\mathrm{SM}(\mathbf{x}) \leq \eta\}dx,
\end{align}
whereas the volume of the dangerous set in a discrete space is given by:
\begin{align}
   \mathrm{vol}(\mathcal{D}) = \int_{\mathbf{x}\in \Omega} \mathbf{1}\{\mathrm{SM}(\mathbf{x}) \leq \eta\}dx.
\end{align}
By comparing the above equations, we can see that the crash rate is actually a weighted sum of the volume metric, where the weights are given by the scenario probabilities.

Now consider the safety comparison between two AVs, denoted as \(Veh_1\) and \(Veh_2\). The difference in crash rates between the two systems is determined by:
\begin{align}
   & P\big(\mathcal{D}(Veh_1)\big)-P\big(\mathcal{D}(Veh_2)\big) \\
   =& \int_{\mathbf{x}\in \Omega} P(\mathbf{x}) \cdot \left[ \mathbf{1}\{\mathrm{SM}(\mathbf{x},Veh_1) \leq \eta\} - \mathbf{1}\{\mathrm{SM}(\mathbf{x},Veh_2) \leq \eta\} \right]dx \\
   =& \int_{\mathbf{x}\in \mathcal{D}(Veh_1, Veh_2)} P(\mathbf{x}) \cdot \left[ \mathbf{1}\{\mathrm{SM}(\mathbf{x},Veh_1) \leq \eta\} - \mathbf{1}\{\mathrm{SM}(\mathbf{x},Veh_2) \leq \eta\} \right]dx,
\end{align}
where \(\mathcal{D}(Veh_1, Veh_2) = \left(\mathcal{D}(Veh_1) \cap \mathcal{S}(Veh_2)\right) \cup \left(\mathcal{D}(Veh_2) \cap \mathcal{S}(Veh_1)\right)\) represents the set of scenarios in which only one of the two AVs is considered unsafe.

Similarly, the difference in volume-based metrics can be expressed as:
\begin{align}
   \mathrm{vol}\big(\mathcal{D}(Veh_1)\big)-\mathrm{vol}\big(\mathcal{D}(Veh_2)\big) = \int_{\mathbf{x}\in \mathcal{D}(Veh_1, Veh_2)} \left[ \mathbf{1}\{\mathrm{SM}(\mathbf{x},Veh_1) \leq \eta\} - \mathbf{1}\{\mathrm{SM}(\mathbf{x},Veh_2) \leq \eta\} \right]dx.
\end{align}
Therefore, the only difference between the two methods lies in whether the scenarios in \(\mathcal{D}(Veh_1, Veh_2)\) are weighted by their occurrence probabilities. If the probabilities of these scenarios differ significantly, then the volume-based and probability-based methods may yield divergent conclusions. However, when the probabilities are approximately uniform, the two methods produce similar results. This condition can be expressed as:
\begin{align}
    P(\mathbf{x}_1) \approx P(\mathbf{x}_2) \approx \text{const}, \quad \forall \mathbf{x}_1 \in \mathcal{D}(Veh_1) \cap \mathcal{S}(Veh_2), \; \mathbf{x}_2 \in \mathcal{D}(Veh_2) \cap \mathcal{S}(Veh_1).
\end{align}

We now analyze under what conditions the approximation between volume-based and probability-based evaluation holds. For the first approximation to be valid, the probabilities of scenarios in \(\mathcal{D}(Veh_1) \cap \mathcal{S}(Veh_2)\) and \(\mathcal{D}(Veh_2) \cap \mathcal{S}(Veh_1)\) must be similar. This typically implies that the safety performance of \(Veh_1\) and \(Veh_2\) is not drastically different. For example, if \(Veh_1\) remains safe in most scenarios while \(Veh_2\) fails in many of them, the set \(\mathcal{D}(Veh_2) \cap \mathcal{S}(Veh_1)\) may contain many high-probability scenarios, violating the approximation. However, in such cases, the set \(\mathcal{D}(Veh_1) \cap \mathcal{S}(Veh_2)\) would likely be much smaller or even empty, and the final comparison would still reflect the true safety ranking.
The second approximation concerns the probability distribution over the set \(\mathcal{D}(Veh_1, Veh_2)\), which typically contains borderline scenarios—those that are risky but avoidable. When both AVs perform well, this set tends to be small and consists mainly of low-probability edge cases, thereby reducing the impact of probability weighting. In such cases, the volume-based method serves as a reliable and data-independent alternative to the probability-based method. 
According to this analysis, when the AVs under comparison exhibit strong safety performance, i.e., have high automation levels, the results from the volume-based and probability-based methods tend to be the same.

\hfill $\square$

Theorem \ref{thm:equivalent} make sure that the volume-based method is accurate for highly automated AVs' full-scenario safety evaluation. By ignoring probability weighting, the volume-based method offers clear advantages in both efficiency and robustness for full-scenario testing. Since it evaluates safety solely based on the volume of dangerous scenarios, it does not require behavior sampling from NDE data, thus avoiding the associated data collection and processing challenges. Moreover, because this method does not rely on scenario probabilities, each scenario has an equal chance of being sampled under Monte Carlo methods, thereby eliminating the curse of rarity.

In the next section, we introduce the algorithms used to compute the scenario volume. For clearer comparison, we focus on the proportion of the dangerous scenario set in the entire driving scenario space, i.e., \(\mathrm{vol}(\mathcal{D})/\mathrm{vol}(\Omega)\), which is consistent with the volume. It is important to note that this ratio is in total different from the crash rate and should not be directly compared with values reported in prior literature.

\section{Volume Calculation for Scenario Sets}
\label{sec:alg}

This section introduces algorithms to calculate the volume of the scenario sets.

\subsection{Monte Carlo Method}
\label{sec:MC}

The Monte Carlo method is a commonly used approach for estimating the volume of high-dimensional and irregular sets. It also serves as a benchmark method in the probability-based method. The proportion of the dangerous scenario set \(\mathcal{D}\) can be estimated by drawing a set of i.i.d. samples \(\mathbf{x}_1, \mathbf{x}_2, \ldots, \mathbf{x}_N\) from the state space \(\Omega\), where \(N\) is the total number of samples. For each sample, the safety metric \(\mathrm{SM}(\mathbf{x})\) is computed through simulation. The volume is then estimated as:
\begin{align}
   \frac{\mathrm{Vol}(\mathcal{D})}{\mathrm{Vol}(\Omega)} = \frac{1}{N} \sum_{i=1}^{N} \mathbf{1}\{\mathrm{SM}(\mathbf{x}_i) \leq \eta\}
\end{align}
where \(\mathbf{1}\{\cdot\}\) denotes the indicator function.

In the Monte Carlo method, we can further define multiple dangerous scenario sets with different risk levels by introducing a set of thresholds \(\{\eta_1, \eta_2, ..., \eta_{E}\}\), where \(E\) corresponds to the highest risk level determined by the safety metric. This allows us to estimate the volume distribution of dangerous scenario sets across different risk levels. Notably, this extension does not increase the computational complexity of the algorithm, as the threshold comparison can be conducted after each simulation run. Algorithm~\ref{alg:mc} summarizes the detailed procedure of the Monte Carlo method using pseudo-code.

\begin{algorithm}[htbp]
   \small
   \renewcommand{\algorithmicrequire}{\textbf{Input:}}
   \renewcommand{\algorithmicensure}{\textbf{Output:}}
   \caption{Monte Carlo estimation for the volume-based evaluation method.}
   \label{alg:mc}
   \begin{algorithmic}[1]
      \REQUIRE AV controller, a set of risk thresholds \(\{\eta_1, \eta_2, ..., \eta_{E}\}\)
      \ENSURE Proportions \(\mathbf{p} = [p_1, p_2, ..., p_E]\) corresponding to each \(\eta_e\)
      \STATE Initialize an array \texttt{Count[$E$]} with zeros
      \FOR{$i = 1$ to $N$}
         \STATE Randomly sample initial parameters of state \(\mathrm{s}\); set \(\mathrm{SM}_{\min} \leftarrow +\infty\)
         \FOR{$t = 0$ to $T$}
            \STATE Randomly sample the behavior of all BVs
            \STATE Determine AV action based on the given controller
            \IF{AV chooses a lane-change maneuver}
               \STATE \textbf{break}
            \ENDIF
            \STATE Update state \(s\); update \(\mathrm{SM}_{\min} \leftarrow \min(\mathrm{SM}_{\min}, \mathrm{SM}(\mathrm{s}))\)
            \IF{a collision occurs}
               \STATE \textbf{break}
            \ENDIF
         \ENDFOR
         \STATE Determine which threshold interval \(\mathrm{SM}_{\min}\) belongs to, and increment corresponding \texttt{Count} entry
      \ENDFOR
      \STATE Compute proportion vector \(\mathbf{p}\) based on \texttt{Count}
      \RETURN \(\mathbf{p}\)
   \end{algorithmic}
\end{algorithm}

Although the Monte Carlo method is applicable to evaluating arbitrary scenarios, its estimation may suffer from high variance and reduced accuracy. Recalling the TTC computation in Equation~\ref{eq:ttc}, TTC is a continuous function of the state variables. In certain scenarios, the scenario set \(\mathcal{D}\) or \(\mathcal{S}\) may form a closed polyhedron. In the next section, we show that for the specific case of a car-following scenario, the volume of the scenario set can be computed analytically using methods for convex sets.

\subsection{Polytope-based Methods}

As mentioned in the last section, we can prove that for a single lane scenario to test AVs' Adaptive Cruise Control function with linear or piecewise linear control law \citep{milanes2014modeling}, the safe scenario set \(\mathcal{D}\) is a convex polyhedron.

\begin{thm}
    \label{thm:convex}
    If the car-following model is linear as shown in Equation \eqref{eq:linear}, the safe scenario set \(\mathcal{S}\) is convex for any threshold \(\eta\).
    \begin{align}
        a^{\mathrm{f}}_{t}=k_1 v^{\mathrm{f}}_{t} + k_2 v^{\mathrm{l}}_{t} + k_3 d_{t} + k_4. \label{eq:linear}
    \end{align}
\end{thm}

\noindent \textit{Proof of Theorem \ref{thm:convex}:}  
By the definition, the safe scenario set can be expressed as
\begin{align}
    \mathcal{S} = \{\mathbf{x} \mid \min\mathrm{TTC}(\mathbf{x}) > \eta\} = \bigcap_{t=0}^{T} \{\mathbf{x} \mid \mathrm{TTC}_t > \eta\}.
\end{align}
For each time step $t$, the condition $\mathrm{TTC}_t > \eta$ can be rewritten as
\begin{align}
    \frac{d_t}{v^{\mathrm{f}}_t-v^{\mathrm{l}}_t} > \eta, \quad v^{\mathrm{f}}_t-v^{\mathrm{l}}_t > 0,
    \quad\Longleftrightarrow\quad
    d_t > \eta (v^{\mathrm{f}}_t-v^{\mathrm{l}}_t), \quad v^{\mathrm{f}}_t-v^{\mathrm{l}}_t > 0.
\end{align}

Since $d_t$, $v^{\mathrm{f}}_t$, and $v^{\mathrm{l}}_t$ are affine functions of $\mathbf{x}$ in the linear car-following model, each condition defines a half-space in $\Omega$. An intersection of half-spaces is convex, thus the safe scenario set $\mathcal{S}$ is convex. 

Specifically, we can express the model using its half-space representation. As stated in the previous section, the decision variables of a scenario include the accelerations \(a^{\mathrm{f}}_t, a^{\mathrm{l}}_t, t\in \mathcal{T}/\{T\}\), speeds \(v^{\mathrm{f}}_t, v^{\mathrm{l}}_t, t\in \mathcal{T}\), and the spatial gap \(d_t, t\in \mathcal{T}\). The feasible set \(\mathcal{S}\) can thus be equivalently represented as the set of all decision variables that satisfy the following linear equality and inequality constraints:
\begin{align}
   &d_{t+1}=d_{t}+\Delta t\, v^{\mathrm{l}}_{t}+\tfrac{1}{2} \Delta t^2\, a^{\mathrm{l}}_{t}-\Delta t\, v^{\mathrm{f}}_{t}-\tfrac{1}{2} \Delta t^2\, a^{\mathrm{f}}_{t}, && \forall t\in \mathcal{T}/\{T\}, \label{eq:dyn_d}\\
   &v^{\mathrm{l}}_{t+1}=v^{\mathrm{l}}_{t}+\Delta t\, a^{\mathrm{l}}_{t}, && \forall t\in \mathcal{T}/\{T\}, \label{eq:dyn_vl}\\
   &v^{\mathrm{f}}_{t+1}=v^{\mathrm{f}}_{t}+\Delta t\, a^{\mathrm{f}}_{t}, && \forall t\in \mathcal{T}/\{T\}, \label{eq:dyn_vf}\\
   &a^{\mathrm{f}}_{t}=k_1 v^{\mathrm{f}}_{t} + k_2 v^{\mathrm{l}}_{t} + k_3 d_{t} + k_4, && \forall t\in \mathcal{T}/\{T\}, \label{eq:cf_model}\\
   &d_{t} - l \geq \eta (v^{\mathrm{f}}_{t}-v^{\mathrm{l}}_{t}), && \forall t\in \mathcal{T}, \label{eq:safe_cond}\\
   &d_{t} - l \geq 0, && \forall t\in \mathcal{T}, \label{eq:safe_cond2}\\
   &d_{\min } \leq d_{0} \leq d_{\max }, \label{eq:dom_d0}\\
   &a_{\min } \leq a^{\mathrm{l}}_{t} \leq a_{\max }, && \forall t\in \mathcal{T}/\{T\}, \label{eq:dom_al}\\
   &a_{\min } \leq a^{\mathrm{f}}_{t} \leq a_{\max }, && \forall t\in \mathcal{T}/\{T\}, \label{eq:dom_af}\\
   &v_{\min } \leq v^{\mathrm{l}}_t \leq v_{\max }, && \forall t\in \mathcal{T}, \label{eq:dom_vl}\\
   &v_{\min } \leq v^{\mathrm{f}}_t \leq v_{\max }, && \forall t\in \mathcal{T}. \label{eq:dom_vf}
\end{align}

Constraints~\eqref{eq:dyn_d}--\eqref{eq:dyn_vf} represent the vehicle dynamics. Constraints~\eqref{eq:cf_model} are a generalized linear car-following model, where $k_1, k_2, k_3$, and $k_4$ are the model's parameters. Constraint~\eqref{eq:safe_cond}--\eqref{eq:safe_cond2} define the safe scenarios, i.e., the TTC equal to or larger than the threshold $\eta$. Constraints~\eqref{eq:dom_d0}--\eqref{eq:dom_vf} specify the domains of the variables, where the parameters \(d_{\min}, d_{\max}, v_{\min}, v_{\max}, a_{\min}, a_{\max}\) define the feasible ranges for spacing, speed, and acceleration. These ranges are determined by vehicle dynamics, road speed limits, and physical constraints. Since all constraints are affine, the feasible set \(\mathcal{D}\) forms a convex polyhedron in the space of decision variables.

\hfill $\square$

Given the convexity established in Theorem~\ref{thm:convex}, both exact and sampling–based strategies based on convex optimization can be applied to evaluate the volume \(\mathrm{Vol}(P)\) of a \(q\)-dimensional polytope \(P\subset\mathbb{R}^{q}\). By computing the volume of the safe scenario set \(\mathrm{Vol}(\mathcal{S})\), defined by the constraints~\eqref{eq:dyn_d}--\eqref{eq:dom_vf}, and the volume of the overall driving scenario space \(\mathrm{Vol}(\Omega)\), defined by constraints~\eqref{eq:dyn_d}, \eqref{eq:cf_model}, and~\eqref{eq:dom_d0}--\eqref{eq:dom_vf}, we can obtain the proportion of safe scenarios.
To compute these volumes in practice, several commonly used toolkits are available for implementing both exact and approximate algorithms. For example, the vertex–enumeration (VE) algorithm \citep{cohen1979two,dyer1983complexity}, a widely adopted exact method, can be efficiently implemented using Python packages such as \texttt{pycddlib} \citep{pycddlib} and \texttt{scipy.spatial.ConvexHull} \citep{emiris2018practical,virtanen2020scipy}. On the other hand, the C++ package \texttt{volesti} \citep{chalkis2020volesti} provides two improved sampling-based heuristic algorithms built upon the Lovász–Vempala algorithm \citep{lovasz2007geometry}. In our experiments, we implement and evaluate all four algorithms. To distinguish them from standard black-box Monte Carlo methods, we refer to these approaches as \textbf{polytope-based methods}, as they explicitly exploit the convex structure of the feasible set. Below, we briefly introduce the main ideas of these algorithms \citep{emiris2018practical,virtanen2020scipy}:

\begin{itemize}
    \item The VE algorithm first enumerates the full vertex set \(\{\mathbf{z}_1,\dots,\mathbf{z}_Z\}\) of the convex polytope \(P = \mathrm{conv}\{\mathbf{z}_1, \dots, \mathbf{z}_Z\}\), where \(\mathbf{z}_i \in \mathbb{R}^p\) denotes the \(i\)-th vertex, and \(Z\) is the total number of vertices. The algorithm then partitions \(P\) into a set of non-overlapping simplices \(\{\Delta_i\}_{i=1}^{m}\), and computes the total volume as the sum of the volumes of these simplices:
    \begin{align}
        \mathrm{Vol}(P) = \sum_{i=1}^{m} \mathrm{Vol}(\Delta_i)
        = \frac{1}{p!} \sum_{i=1}^{m} \left| \det\left(\mathbf{z}_{i,1} - \mathbf{z}_{i,0}, \dots, \mathbf{z}_{i,p} - \mathbf{z}_{i,0} \right) \right|,
    \end{align}
    where each simplex \(\Delta_i = \mathrm{conv}\{\mathbf{z}_{i,0}, \dots, \mathbf{z}_{i,p}\}\) is formed by selecting \(p+1\) affinely independent vertices from the set. Due to the exponential growth of the number of vertices \(Z\) with the dimension \(p\), the VE algorithm is generally limited to low-dimensional polytopes.
    
    \item The Sequence of Balls (SOB) algorithm estimates the volume of a convex polytope \(P\) by embedding it between a sequence of concentric Euclidean balls \(B(r_0) \subseteq P \subseteq B(r_R)\) with increasing radii \(r_0 < r_1 < \dots < r_R\), where $R$ is the number of balls. The volume is then approximated using the following telescoping product:
    \begin{align}
        \mathrm{Vol}(P) &= \mathrm{Vol}(B(r_0)) \prod_{i=1}^{R}
        \frac{\mathrm{Vol}(P \cap B(r_i))}{\mathrm{Vol}(P \cap B(r_{i-1}))}, \label{eq:sob1} \\
        \mathrm{Vol}(B(r_0)) &= c_p \, r_0^p, \label{eq:sob2}
    \end{align}
    where \(c_p\) is the volume of the unit ball in \(\mathbb{R}^p\). Each ratio in the product is estimated by drawing random samples from the intersection \(P \cap B(r_i)\) using the Hit-and-Run random walk with relative tolerance \(\varepsilon\). The simplicity and scalability of this method make it particularly effective for moderate dimensions, typically when \(p \lesssim 30\).
    
    \item The Cooling Gaussians (CG) algorithm replaces the geometric sequence of balls with a sequence of Gaussian densities \(g_s(x) \propto \exp\left(-\|x\|^2 / 2\sigma_s^2\right)\), where \(s = 0, 1, \dots, S\) indexes the cooling stages, and the variances follow a geometric cooling schedule \(\sigma_{s+1} = \gamma \sigma_s\) for some fixed \(0 < \gamma < 1\). Let \(Z_s = \int_{P} g_s(x) \, dx\) denote the normalizing constant of the truncated density over the polytope \(P\). Then the volume of \(P\) can be expressed as:
    \begin{align}
        \mathrm{Vol}(P) = Z_0 \prod_{s=0}^{S-1} \frac{Z_{s+1}}{Z_s}. \label{eq:cg}
    \end{align}
    Each ratio \(Z_{s+1}/Z_s\) is estimated as an expectation under the distribution \(g_s(x) \mid_{P}\), using Hit-and-Run samples with walk length \(L\) and convergence tolerance \(\varepsilon\).  This approach delivers high-accuracy volume estimates even in high dimensions (e.g., \(p > 30\)), although it generally requires longer Markov chains and more samples compared to other methods.
    
\end{itemize}

When computing the volume, we directly employ the half-space representation defined by Constraints~\eqref{eq:dyn_d}--\eqref{eq:dom_vf}. Although the original model involves \(5T\) decision variables, the presence of equality constraints—specifically the vehicle dynamics and car-following equations—implies that the feasible set lies within a lower-dimensional affine subspace. In particular, these equality constraints reduce the effective dimensionality of the feasible region to \(T + 3\). To exploit this structure, we apply a null-space projection \citep{byrd1986continuity} to eliminate the equality constraints and project the problem into a lower-dimensional subspace. Specifically, we compute a basis for the null space of the equality constraint matrix and express all feasible solutions as linear combinations of this basis and a particular solution that satisfies the equalities. This transformation significantly reduces the dimensionality of the volume computation problem, enabling more efficient evaluation using either exact or sampling-based algorithms.

\section{Experimental Results}
\label{sec:exp}

\subsection{Experiment Setting}

The experiment is conducted in both single-lane and multi-lane scenarios to test the Monte Carlo method and polytope-based methods. In the single-lane scenario, the AV serves as the FV. In the multi-lane scenario, we consider four BVs around the AV. Since the length of the testing scenario is a key parameter in our method, we test different lengths in \( T \) with a time interval \( \Delta t = 0.2 \) seconds. In the single-lane scenario, we set the range as \(d_{\min}=5~\text{m}, d_{\max}=100~\text{m}, v_{\min}=0~\text{m/s}, v_{\max}=40~\text{m/s}, a_{\min}=-4~\text{m/s}^2, a_{\max}=2~\text{m/s}^2\). In the multi-lane scenario, the AV also needs to make decisions on left and right lane changes.

For the behavior models of the tested AV, we consider two representative car-following models: the linear car-following model proposed by \citet{milanes2014modeling}
The linear car-following model is formulated as:
\begin{align}
    a_t = k_1 \left(x^{\mathrm{l}}_{t} - x^{\mathrm{f}}_t - t_{\mathrm{hw}} v^{\mathrm{f}}_t \right) + k_2 \left(v^{\mathrm{l}}_{t} - v^{\mathrm{f}}_t\right), \label{eq:linear_cf}
\end{align}
where \(t_{\mathrm{hw}}\) is the desired time headway setting, and \(k_1\) and \(k_2\) are control gains on the position and speed errors, respectively. In our implementation, we use the parameters from \citet{milanes2014modeling}: \(k_1 = 0.23~\mathrm{s}^{-2}\) and \(k_2 = 0.07~\mathrm{s}^{-1}\), and set \(t_{\mathrm{hw}}=1.5~\mathrm{s}\).

In multi-lane scenarios, we adopt the Minimizing Overall Braking Induced by Lane changes (MOBIL) model \citep{treiber2016mobil} to describe lane-changing behavior. MOBIL is a utility-based model that evaluates whether a lane change should be performed based on the trade-off between the advantage to the ego vehicle and the disadvantage (i.e., braking) imposed on surrounding vehicles. The decision criterion is defined by the incentive function:
\begin{align}
    \Delta a_{\mathrm{ego}} + f \left( \Delta a_{\mathrm{new}} + \Delta a_{\mathrm{old}} \right) > a_{\mathrm{th}}, \label{eq:mobil}
\end{align}
where \(\Delta a_{\mathrm{ego}}\) is the expected acceleration gain for the ego vehicle after the lane change. \(\Delta a_{\mathrm{new}}\) and \(\Delta a_{\mathrm{old}}\) are the expected changes in acceleration for the new follower and the old follower, respectively. \(f\) is the politeness factor (\(0 \leq f \leq 1\)), representing the weight the ego vehicle places on other drivers’ comfort. \(a_{\mathrm{th}}\) is the minimum incentive threshold to trigger a lane change. A lane change is allowed only if the incentive criterion in Equation~\eqref{eq:mobil} is satisfied and the resulting deceleration of the new follower does not exceed a safety threshold. This ensures that the lane change benefits the ego vehicle without causing excessive braking for nearby vehicles. In our experiment, we set $f=0$.

To validate the volume-based evaluation method, we apply it to commercially available production AVs. The behavior models of these AVs are calibrated using field-test trajectories collected from the Ultra-AV dataset \citep{zhou2024unified}, which contains data from over 30 types of ACC-equipped vehicles. We randomly selected from the Ultra-AV dataset \cite{zhou2024unified} and denoted as \( Veh_A \) to \( Veh_F \). Since there are currently very few commercial AVs capable of autonomous lane-changing and limited corresponding trajectory datasets, we only test these AVs in single-lane scenarios. The car-following models of each AV are calibrated using the linear model in Equation \eqref{eq:cf_model}. Table \ref{tab:calibration} shows the calibration results, where the root mean square error (RMSE) is listed to show the accuracy of the car-following model.

\begin{table}[htbp]
  \centering
  \caption{Calibration results of the car-following models for $Veh_A$-$Veh_F$.}
    \begin{tabular}{lrrrrrr}
    \toprule
    Vehicle ID & $Veh_A$ & $Veh_B$ & $Veh_C$ & $Veh_D$ & $Veh_E$ & $Veh_F$ \\
    \midrule
    RMSE (m/s$^2$)  & 0.121 & 0.342 & 0.272 & 0.336 & 0.346 & 0.256 \\
    $k_1$ (s$^{-2}$)   & 0.018 & 0.004 & 0.001 & 0.006 & 0.003 & 0.012 \\
    $k_2$ (s$^{-1}$)    & 0.156 & 0.241 & 0.308 & 0.249 & 0.257 & 0.168 \\
    \(t_{\mathrm{hw}}\) (s)   & 1.378 & 2.379 & 0.467 & 2.002 & 2.225 & 2.424 \\
    \bottomrule
    \end{tabular}%
  \label{tab:calibration}%
\end{table}%

The Monte Carlo and VE algorithms were implemented in Python. The SOB and CG algorithms were executed via Python bindings to the C++ package \texttt{volesti}. The maximum number of iterations for the Monte Carlo method was set to 1,000,000. For SOB and CG, the relative tolerance \(\varepsilon\) was set to 0.01, and the walk length $L$ for CG was set to 1,000. All experiments were conducted on a machine equipped with a 3.2 GHz AMD Ryzen 7 7735HS CPU with Radeon Graphics and 16 GB of RAM, running the Ubuntu 22.04 operating system. \textbf{Code and data will be made publicly available upon acceptance of this paper.}

\subsection{Performance of the Volume Estimation Algorithms in the Single-lane Scenario}

We first evaluate the performance of the four algorithms in a single-lane scenario. Since the dimensionality of the problem increases with the time horizon \(T\), we test the algorithms under various dimensions with \(T \in \{1, 2, 3, 4, 5, 10, 15, 20, 25\}\). Here, \(T = 25\) corresponds to a trajectory length of 5 seconds, which is consistent with the typical scenario duration in the literature. In this section, we use \textbf{MC} to denote the Monte Carlo method. In the results table, columns \textbf{Percent} report the proportion of dangerous scenarios, columns \textbf{Time} report the computation time of each algorithm in seconds, and columns \textbf{Error} report the percentage error of each algorithms' dangerous scenario proportion compared with the exact results of the VE algorithm. Here we set $\eta=1$~s, which means that the dangerous scenarios have \(\mathrm{TTC} \leq 1\). A maximum runtime limit of 3,600 seconds was set on all algorithms. For algorithms that exceed the predefined time limit, the results are marked with \textbackslash.

\begin{table}[htbp]
  \centering
  \small
  \caption{Comparison between the Monte Carlo method and the polytope-based methods.}
    \begin{tabular}{crrrrrrrrrrrr}
    \toprule
    \multirow{2}[4]{*}{T} & \multicolumn{3}{c}{VE} & \multicolumn{3}{c}{SOB} & \multicolumn{3}{c}{CG} & \multicolumn{3}{c}{MC} \\
    \cmidrule{2-13}          & Percent & Time  & Error & Percent & Time  & Error & Percent & Time  & Error & Percent & Time  & Error \\
    \midrule
    1     & 3.59  & 0.08  & \textbackslash{} & 3.90  & 53.28 & 8.71  & 4.50  & 39.83 & 25.48 & 3.63  & 10.46 & 1.06 \\
    2     & 4.21  & 0.47  & \textbackslash{} & 4.58  & 119.02 & 8.74  & 5.19  & 62.66 & 23.28 & 4.22  & 12.63 & 0.24 \\
    3     & 4.90  & 3.17  & \textbackslash{} & 4.44  & 218.12 & 9.36  & 6.21  & 84.21 & 26.64 & 4.93  & 14.48 & 0.62 \\
    4     & 5.69  & 58.18 & \textbackslash{} & 6.19  & 328.82 & 8.79  & 7.07  & 104.14 & 24.18 & 5.74  & 16.52 & 0.76 \\
    5     & 6.77  & 2162.00 & \textbackslash{} & 6.31  & 1242.51 & 6.77  & 8.93  & 118.07 & 31.94 & 6.73  & 18.33 & 0.54 \\
    10    & \textbackslash{} & \textbackslash{} & \textbackslash{} & \textbackslash{} & \textbackslash{} & \textbackslash{} & 10.65 & 291.87 & \textbackslash{} & 11.24 & 26.53 & \textbackslash{} \\
    15    & \textbackslash{} & \textbackslash{} & \textbackslash{} & \textbackslash{} & \textbackslash{} & \textbackslash{} & 12.59 & 754.29 & \textbackslash{} & 14.06 & 32.93 & \textbackslash{} \\
    20    & \textbackslash{} & \textbackslash{} & \textbackslash{} & \textbackslash{} & \textbackslash{} & \textbackslash{} & 15.78 & 1433.86 & \textbackslash{} & 16.11 & 52.14 & \textbackslash{} \\
    25    & \textbackslash{} & \textbackslash{} & \textbackslash{} & \textbackslash{} & \textbackslash{} & \textbackslash{} & 19.04 & 1797.84 & \textbackslash{} & 17.21 & 46.51 & \textbackslash{} \\
    \bottomrule
    \end{tabular}%
  \label{tab:single}%
\end{table}%

\begin{figure}
    \centering
    \includegraphics[width=0.75\linewidth]{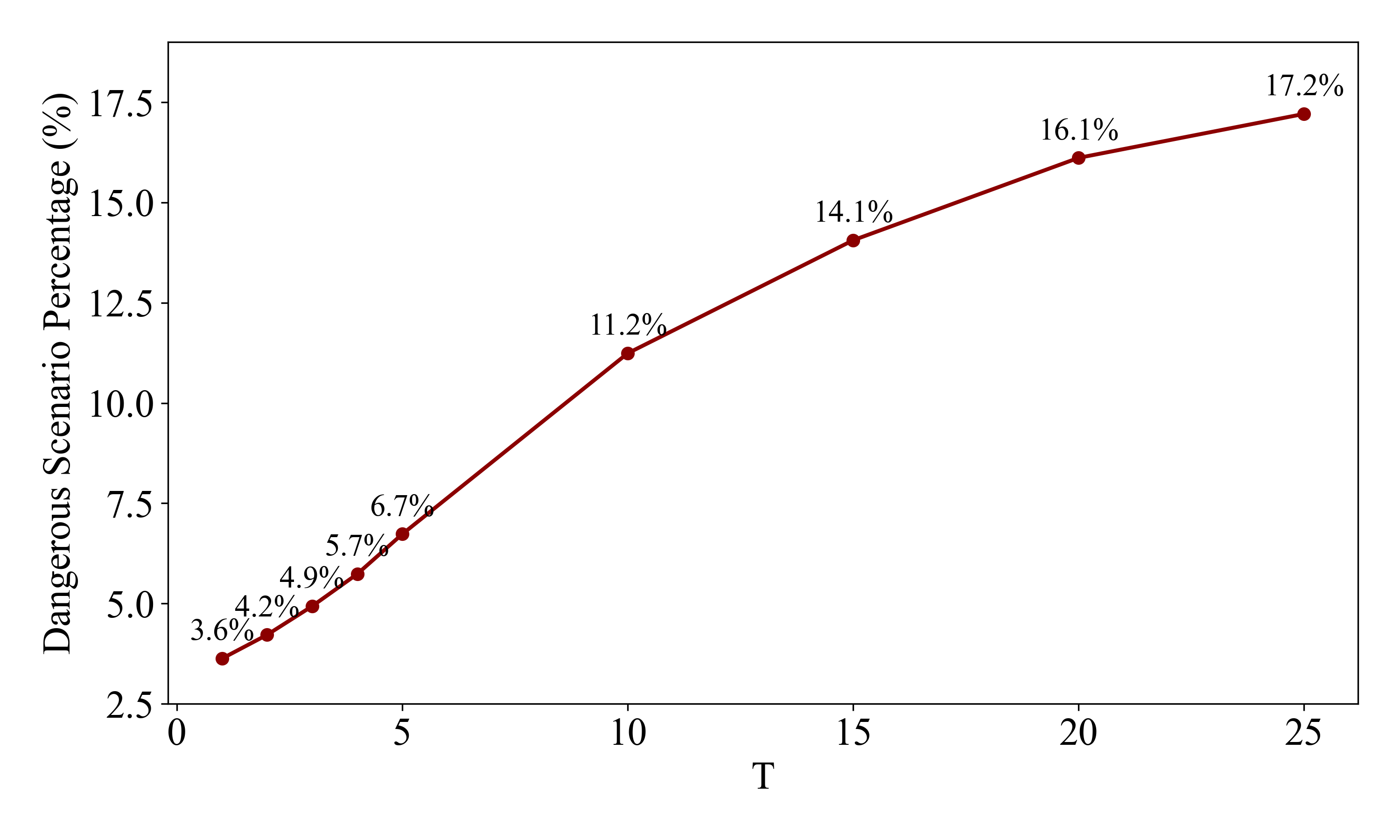}
    \caption{Trend of the proportion of dangerous scenarios identified by the MC method.}
    \label{fig:dangerous}
\end{figure}

Table~\ref{tab:single} presents the performance of the four algorithms under different time horizons. In terms of the proportion of dangerous scenarios, all algorithms exhibit a consistent trend: as the time horizon \(T\) increases, the proportion of dangerous scenarios also increases. This trend is further visualized in Figure~\ref{fig:dangerous}, which shows a line plot of the dangerous scenario proportion computed by the MC algorithm for \(T = 1\) to \(T = 25\). The curve displays a shape similar to a convex function. This observation is expected, as longer scenarios inherently include shorter ones—meaning that any dangerous scenario identified under a smaller \(T\) will also be included in the count for larger \(T\) values.

Regarding computational time, the exact VE algorithm exhibits exponential growth for \(T\) and becomes intractable when \(T > 5\). Therefore, VE primarily serves as a benchmark to validate the results of other algorithms. Among the heuristic methods, the SOB algorithm also fails to return results within the time limit for \(T = 10\). In contrast, both CG and MC remain computationally feasible even for longer horizons. The MC method in particular shows polynomial growth for \(T\), making it a scalable option for larger problems.

Regarding accuracy, the MC method demonstrates high precision, with errors within 1\% for small values of \(T\) compared to the exact VE results. Surprisingly, although both SOB and CG are designed specifically for convex polytopes and leverage geometric structure, their performance is inferior to the basic MC method. The observed errors are between 5–10\% for SOB and 20–30\% for CG, which are significantly higher than those of MC. This may be because although the safe set is convex, its geometry can be irregular, potentially undermining the effectiveness of structure-aware sampling strategies.

Overall, the Monte Carlo method appears to be the most favorable option for volume-based evaluation. It offers relatively short computation time, scales well with longer horizons, and achieves high accuracy compared to the exact VE method.

\subsection{Scenario Distributions in Single-lane and Multi-lane Scenarios}

In this section, we further evaluate the effectiveness of the proposed volume-based evaluation method in multi-lane scenarios. Since the Monte Carlo method demonstrated strong performance in the previous section, we focus exclusively on its results in this analysis. As mentioned in Section~\ref{sec:MC}, the Monte Carlo method enables the estimation of scenario distributions across different levels of risk. Accordingly, we divide the scenario space into several blocks: one block corresponds to crash scenarios, another corresponds to dangerous scenarios with \(\mathrm{TTC} \in [0, 5]\) divided into intervals of 0.5, and the remaining block consists of safe scenarios where \(\mathrm{TTC} > 5\). To further assess the accuracy and sensitivity of the volume-based evaluation method, we test three car-following behavior models with different desired time-gap settings, \(t_{\mathrm{hw}} \in \{1.0, 1.5, 2.0\}\) seconds, while keeping all other parameters fixed. Under normal circumstances, a smaller time-gap setting is expected to correspond to a higher level of driving risk.

\begin{figure}
    \centering
    \includegraphics[width=0.75\linewidth]{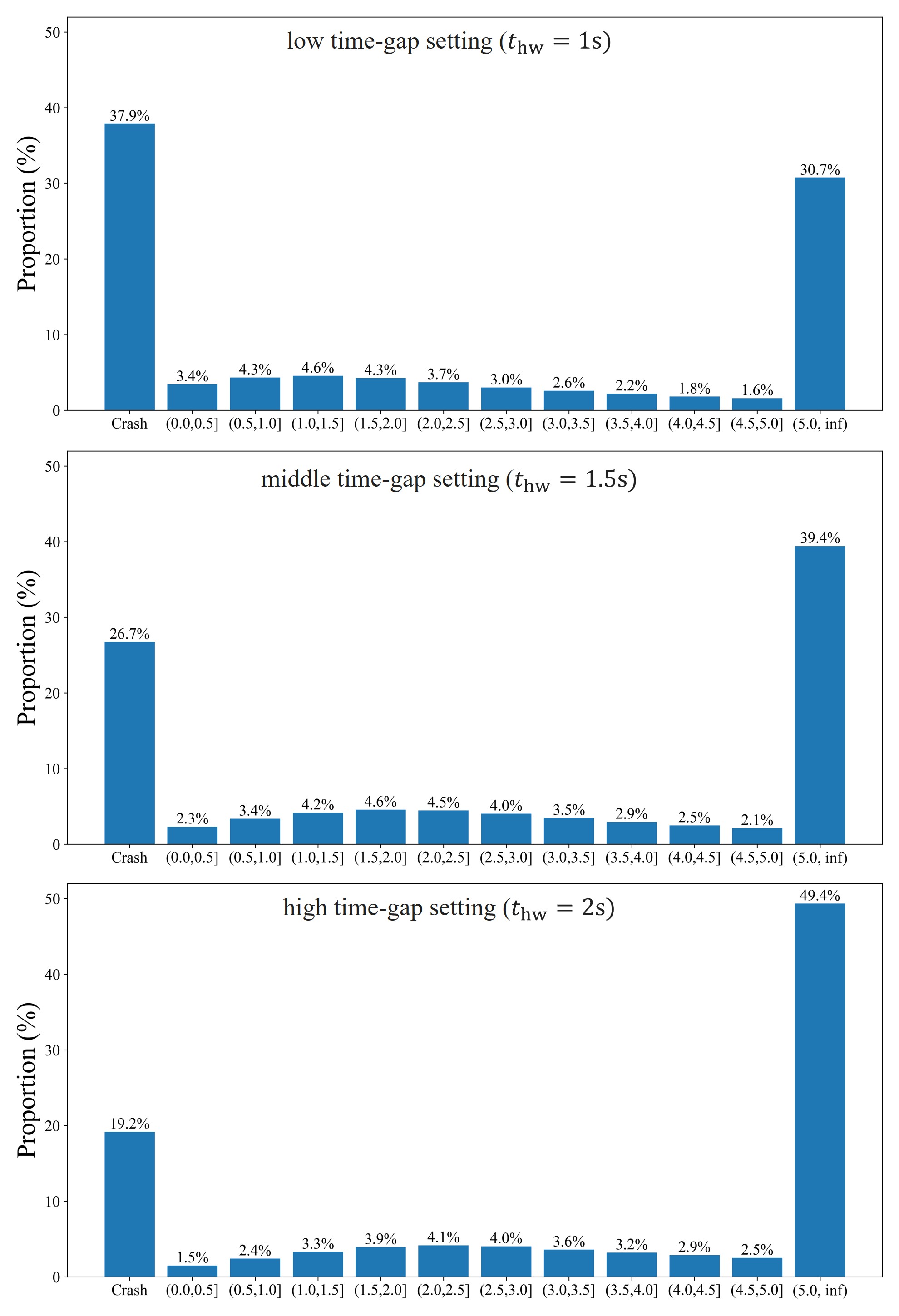}
    \caption{Distribution of scenario risk levels for three car-following models in the single-lane scenario.}
    \label{fig:cf}
\end{figure}

\begin{figure}
    \centering
    \includegraphics[width=0.75\linewidth]{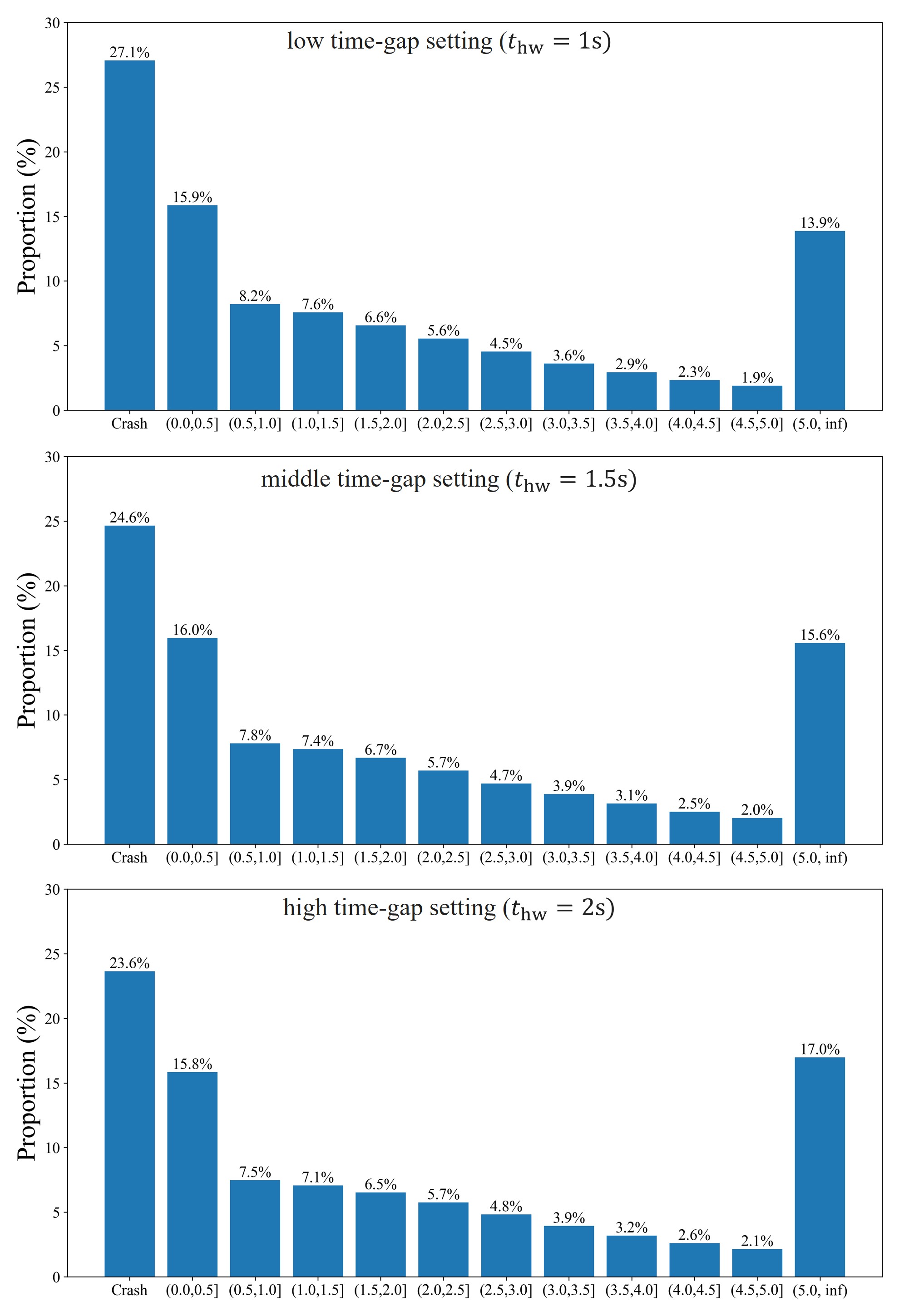}
    \caption{Distribution of scenario risk levels for three car-following models in the multi-lane scenario.}
    \label{fig:lc}
\end{figure}

Figures~\ref{fig:cf}--\ref{fig:lc} show the distribution results for the single-lane and multi-lane scenarios, respectively. In each figure, the horizontal axis represents different scenario categories, while the vertical axis indicates the proportion of each category within the total set of scenarios. First, we observe a clear trend in both figures: as the desired time-gap increases, the proportion of crash scenarios decreases, while the proportion of safe scenarios (with \(\mathrm{TTC} > 5\)) increases. Notably, in the single-lane scenario, the distribution of dangerous scenarios with \(\mathrm{TTC} \in [0, 5]\) shifts from left-skewed to right-skewed as the time-gap increases. Although this shift is less pronounced in the multi-lane scenario, there is still a noticeable reduction in the proportion of lower-TTC scenarios and a corresponding increase in higher-TTC scenarios. These observations confirm that the volume-based evaluation method can effectively capture the impact of the time-gap setting on safety performance. Second, we find that the distributions differ significantly between single-lane and multi-lane scenarios. In the single-lane case, the majority of scenarios fall into the safe category (\(\mathrm{TTC} > 5\)), while in the multi-lane case, most scenarios are concentrated in the dangerous region with \(\mathrm{TTC} < 5\), and exhibit a roughly decreasing trend as the danger level increases. This highlights the fact that multi-lane driving tends to be substantially more hazardous due to the added complexity and interactions among vehicles. Overall, the distribution patterns are consistent with common driving safety intuitions, indicating that the volume-based evaluation method yields reliable and interpretable results.

\subsection{Applying the Volume-based Evaluation Method in Production AVs}
\label{sec:pav}

In this section, we apply the proposed volume-based evaluation method to the testing of production AVs. The Monte Carlo method is used to evaluate six car-following models calibrated from AV trajectory data, with the time horizon set to \(T = 25\).

\begin{figure}
    \centering
    \includegraphics[width=1.\linewidth]{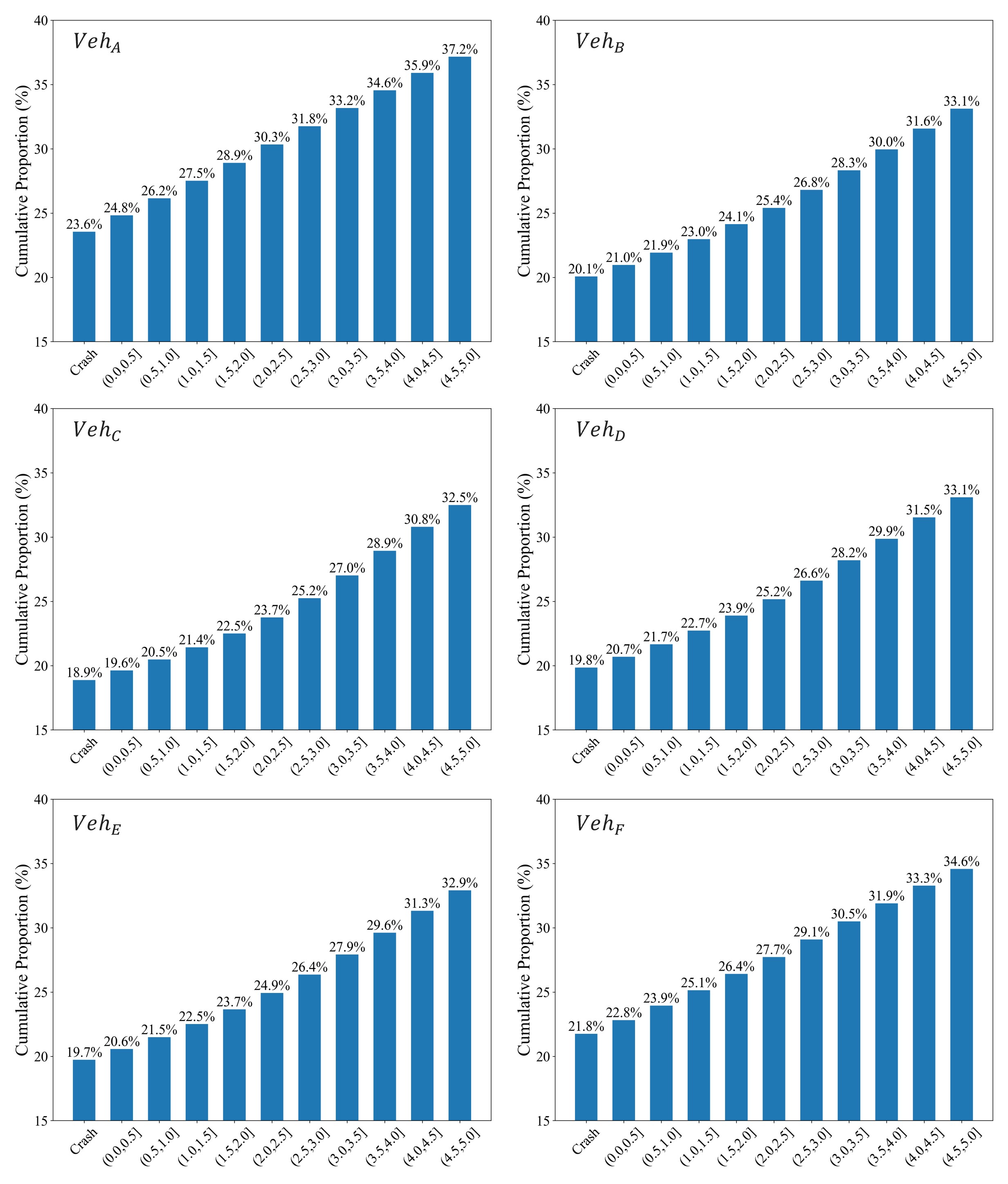}
    \caption{Comparison of the distributions of production AVs.}
    \label{fig:pav}
\end{figure}

Figure~\ref{fig:pav} shows the cumulative proportion distribution of the six tested AVs. The horizontal axis is consistent with Figures~\ref{fig:cf}--\ref{fig:lc}, but for clarity, safe scenarios are omitted. The vertical axis represents the cumulative proportion of scenarios at or below a given safety level, i.e., scenarios that are equally or more dangerous. The figure clearly illustrates significant differences in the proportion of dangerous scenarios across vehicles. For example, $Veh_A$ exhibits a noticeably higher proportion of dangerous scenarios, especially crash cases, compared to other AVs. In contrast, $Veh_C$ shows the lowest proportion of dangerous scenarios, indicating superior safety performance. Based on the proportion of dangerous scenarios, the safety ranking of the tested AVs can be roughly ordered as $Veh_C$, $Veh_E$, $Veh_D$, $Veh_B$, $Veh_F$, and $Veh_A$. The distributions show a generally dominating pattern, supporting the reasonableness of the safety ranking.

Interestingly, $Veh_C$—the safest vehicle according to this evaluation—uses a car-following model with the smallest desired time gap, as reported in Table~\ref{tab:calibration}. This contradicts the common intuition that a shorter following distance implies higher risk. However, $Veh_C$'s controller has a relatively large value for the parameter \(k_2\), indicating higher sensitivity to the speed difference between the ego vehicle and the leading vehicle. Since the speed difference appears in the denominator of the TTC formula, this sensitivity has a strong influence on TTC and may partially explain the high proportion of safe scenarios for $Veh_C$. This result provides an important insight for AV developers: although there is often a trade-off between mobility and safety, it is possible to achieve both through better controller design or by improving hardware-level performance, such as reducing communication delay.

\section{Conclusion} 
\label{sec:con}

This paper proposes a novel framework for evaluating the safety of AVs under diverse and complex driving scenarios. This method can address the limitations of traditional probability-based evaluation methods, which strongly depend on naturalistic data and are vulnerable to the curse of dimensionality. The proposed modeling framework standardizes diverse traffic scenarios into a structured representation involving three lanes and six surrounding background vehicles. This compact model effectively captures key interactions while keeping the dimensionality tractable. To quantify AV safety without relying on scenario probabilities, we define a volume-based metric that measures the proportion of risky scenarios within the total scenario space. For car-following scenarios, we prove that under a linear behavior model and TTC-based safety definition, the safe scenario set is convex. This property enables the application of polytope volume computation algorithms, such as vertex enumeration and sampling-based methods. Experimental results demonstrate that the Monte Carlo method provides both accurate and scalable volume estimates, while heuristic polytope-based methods offer complementary insights. Furthermore, we validate the proposed framework using six production AVs calibrated from field-test trajectory data, and show that the evaluation results align with intuitive safety expectations and controller parameter settings.

This study provides an interpretable, data-efficient, and extensible framework for AV safety evaluation. Although the unified scenario model is primarily used in the volume-based method in this paper, it is not limited to this context and can also be integrated into conventional AV crash risk estimation approaches. Future work may explore such integration and compare the results against those obtained from volume-based evaluation. Due to the limited availability of behavior models and driving data in the current literature, we were not able to evaluate the framework across a broader range of scenarios. As a next step, the proposed method could be implemented in simulators such as CARLA \citep{dosovitskiy2017carla}, enabling evaluation of end-to-end learning-based controllers under more diverse and controllable conditions. Furthermore, the convexity result established for car-following scenarios may potentially be extended to other types of driving scenarios. However, such extensions may require the use of alternative safety metrics that capture lateral interactions or the influence of vehicles in adjacent lanes. Exploring more general conditions under which the safe scenario set remains convex is an important direction for future theoretical work. Finally, a promising line of research is to integrate the insights from safety evaluation into AV controller design, enabling the development of autonomous systems that are not only performant but also verifiably safe under a wide range of real-world scenarios.

\section{Acknowledgement}

This research was sponsored by the National Science Foundation, CPS: Small: NSF-DST: Turning “Tragedy of the Commons (ToC)” into “Emergent Cooperative Behavior (ECB)” for Automated Vehicles at Intersections with Meta-Learning (No. 2343167).

\bibliographystyle{informs2014}
\bibliography{reference.bib}

\end{document}